\documentclass[letterpaper, 10 pt, conference]{ieeeconf}  

\IEEEoverridecommandlockouts                              

\usepackage[noadjust]{cite}
\usepackage{booktabs}
\usepackage{float}
\usepackage{amsmath}
\usepackage{amsfonts}
\usepackage{graphicx}
\usepackage{subcaption}
\let\labelindent\relax
\usepackage{enumitem}
\usepackage[dvipsnames,table]{xcolor}
\usepackage{textcomp, gensymb}
\usepackage{hyperref}
\usepackage{lipsum}
\usepackage{balance}
\usepackage[switch]{lineno}
\usepackage{mathtools}
\usepackage[linesnumbered,ruled,vlined,english,onelanguage]{algorithm2e}

\newcommand{\hyper}[1]{\textcolor{magenta}{\texttt{#1}}}

\definecolor{pred_red}{HTML}{EF5948}
\definecolor{pred_yellow}{HTML}{F3B116}
\definecolor{pred_blue}{HTML}{2B9BE7}
\definecolor{pred_green}{HTML}{03B56E}

\title{\LARGE \bf
Simultaneous Localization and Affordance Prediction\\ of Tasks from Egocentric Video
}

\author{Zach Chavis$^{1}$, Hyun Soo Park$^{1}$, and Stephen J. Guy$^{1}$\\
{\tt\small\hyper{https://appliedmotionlab.github.io/slap}}
\thanks{$^{1}$Department of Computer Science and Engineering (CS\&E), University of Minnesota, Minneapolis, MN 55414 US.
{\tt\small(chavi014|hspark|sjguy)@umn.edu}}%
}

\let\oldtwocolumn\twocolumn
\renewcommand\twocolumn[1][]{%
    \oldtwocolumn[{#1}{
           \centering
           \includegraphics[width=0.8\textwidth]{figures/teaser.pdf}
           \captionof{figure}{\small{\textbf{Predicting Spatial Task Affordance.} \textbf{(left)}~Existing vision and language models (VLMs) are able to localize seen objects related to tasks within an image (colored regions) or reason about tasks through vision-language similarity (colored bars). However, existing VLMs are unable to predict into the 3D space outside of the given image itself. \textbf{(right)}~We propose augmenting VLMs to make spatial predictions for where given tasks likely take place relative to an egocentric image. We refer to this region of a task's likely locations as the task's spatial affordance.} }
           \label{fig:teaser}
    }]
}

\begin{document}

\maketitle

\begin{abstract}

Vision-Language Models (VLMs) have shown great success as foundational models for downstream vision and natural language applications in a variety of domains. However, these models are limited to reasoning over objects and actions currently visible on the image plane. We present a spatial extension to the VLM, which leverages spatially-localized egocentric video demonstrations to augment VLMs in two ways --- through understanding \textit{spatial task-affordances}, i.e. where an agent must be for the task to physically take place, and the localization of that task relative to the egocentric viewer. We show our approach outperforms the baseline of using a VLM to map similarity of a task's description over a set of location-tagged images. Our approach has less error both on predicting where a task may take place and on predicting what tasks are likely to happen at the current location. The resulting representation will enable robots to use egocentric sensing to navigate to, or around, physical regions of interest for novel tasks specified in natural language.

\end{abstract}

\section{INTRODUCTION}
Understanding spatial affordances, i.e., the region of space in which a task can be accomplished in an environment, is a vital capability in any robotic or AI system that seeks to model or imitate how humans use the environment around them. Such affordances may be naturally learned from human demonstrations. Egocentric video demonstrations, i.e. first-person video captured from a head-mounted camera, is especially well-suited for learning spatial affordances as it simultaneously captures where a person is going and what they are seeing and using as they move through their environment~\cite{plizzari2024outlook}. In particular, recent large data collection efforts such as Ego4D~\cite{Grauman_2022_CVPR} and EgoExo4D~\cite{egoexo4d} provide high quality egocentric data with each frame localized in space together with annotations capturing  narrative descriptions of the tasks being accomplished at each stage in the video.

Recent work has proposed systems that can provide high-quality reasoning about object's affordances, that is how objects can be used for tasks. For example, CLIP-Fields~\cite{clipfields} allow robots to reason over semantic maps to find 3D locations of objects for tasks such as identifying the location of a microwave when given the task ``warm up my lunch.'' However, these kinds of systems rely on access to a full 3D model at inference time (e.g., through a NeRF~\cite{nerfs} or 3D point cloud).

Here, we define a task's spatial affordance as the area in free space where a person would stand in order to perform the task. This type of knowledge is important for robots in human environments, as it will help them better understand where people will likely be as they are doing tasks. We consider the problem of predicting 3D regions of spatial affordances from a single, egocentric image. 

We conceptualize this problem as first understanding the scene context from an image, and then combining this context with a given task in order to predict the likely region where a person may be. We propose a neural-network based approach which solves both problems simultaneously with an encoder-decoder style architecture. The resulting network is trained on a large set of tasks from a variety of cooking activities and kitchen environments and is able to predict new spatial task affordances given natural language descriptions.

Our problem is closely related to the use of Vision-Language Models (VLMs) in robotics. Figure \ref{fig:teaser}(left), outlines common uses of VLMs in robotics, such as segmenting objects a robot may interact with for its task~\cite{paligemma, owl}, or measuring the similarity of a current ego image with a task the robot is interested in~\cite{clip,nair2022r3m}. In contrast, our proposed framework provides a new capability: Given a single egocentric image, rather than identifying items or measuring similarities, our model produces spatial 3D regions of task's location Figure~\ref{fig:teaser}(right).

When deployed to new tasks and views in known environments (seen in training), our resulting system outperforms baselines, even when baselines are provided with many images (entire demonstration) at inference time rather than the single image used in our approach. We build on the proposed spatial task affordance predictions to introduce the concept of a task obstacle. These are regions over sets of related tasks which can be used to guide robot avoidance in human environments.

In summary, our main contributions are:
\begin{itemize}
    \item An extension of VLMs to predict 3D spatial regions representing task location likelihood.
    \item Using the EgoExo4D dataset to learn spatial affordances from egocentric human demonstrations in real kitchen environments.
    \item A training approach which allows for optimizing a model on spatial demonstrations of tasks on views across the full environment.
    \item Task obstacles to enable robots to avoid potential collisions in human environments.
\end{itemize}

\section{RELATED WORK}

Deep learning has proven to be a powerful paradigm for understanding scene geometry from images, both in multi-image scene reconstruction as seen in NeRFs~\cite{nerfs}, and single-frame third-person body pose prediction~\cite{caoHMP2020}, first-person navigation~\cite{park_cvpr_future_loc:2016,Qiu2021EgocentricHT,Pan2022COPILOTHC}, and first-person body pose prediction~\cite{wang2023scene} tasks. Beyond geometry, semantic reasoning through natural language over images has recently been enabled via Vision-Language Models (VLM) such as CLIP~\cite{clip}, BLIP~\cite{li2022blip}, and EgoVLP~\cite{egovlp2022}. However, these models on their own have limited spatial understanding~\cite{elbanani2024probing}.

Egocentric vision is a common representation for robots due to the prevalence of on-board cameras. As such, methods have been developed to leverage egocentric data for robotic tasks such as identifying activities~\cite{egoactivity}, shaping behavior~\cite{Nagarajan2021ShapingEA}, and inferring goal locations~\cite{Datta2020IntegratingEL}. To support these applications, specialized large-scale datasets of egocentric human demonstrations have been proposed, such as the Ego4D dataset~\cite{Grauman_2022_CVPR}, and the EgoExo4D dataset~\cite{egoexo4d}.

Recent work seeks to align geometry and semantics to enable robust navigation of mobile agents. Reinforcement learning approaches seek to understand how to reason about the environment given a pre-defined task from a robot's perspective~\cite{pointnav, thda, hutterwalk, Partsey_2022_CVPR, navobj, pmlr-v100-kumar20a, shah2022lmnav}. Vision-Language Navigation approaches~\cite{reverie} seek enable robot navigation in human environments, but focus on objects as opposed to tasks. CLIP has been integrated into mobile robot policies to allow natural language task augmentation~\cite{majumdar2022zson,Dorbala2022CLIPNavUC,shah2022lmnav}. VLMs have also been used to create flexible semantic maps a mobile robot can query using natural language, such as VLMaps~\cite{vlmaps}, NLMap-SayCan~\cite{nlmapsaycan}, CLIP-Fields~\cite{clipfields}, and 3D-LLMs~\cite{3dllm}, using e.g. an RRT~\cite{karaman2011sampling}.

A closely related problem to spatial affordances (where a person stands for a task) is manipulation affordances (how to manipulate an object for a task). Manipulation affordances  can be estimated from image segmentation~\cite{paligemma, AffordanceNet18}, from 3D object or scene representations~\cite{wen2021catgrasp,Makhal2017ReuleauxRB}, or learned end-to-end~\cite{ma2023eureka}. Affordances can also be learned from human demonstration as in the Vision-Robotics Bridge~\cite{bahl2023affordances} and its text-based extension~\cite{yoshida2024textdriven} which learn to represent image-based affordances from egocentric human demonstrations, where affordance is defined as contact points and trajectories for robots to interact. R3M~\cite{nair2022r3m} uses egocentric human demonstrations to create a semantic representation well-suited as a foundational model for downstream robot tasks.

\section{SIMULTANEOUS LOCALIZATION AND AFFORDANCE PREDICTION}\label{sec:SLAP}

Given an egocentric image and a natural language distribution of a task (e.g., ``turn on the stove''), our goal is to predict the region where someone would likely be when performing this task. 
We assume that tasks, images, and viewpoints are new and unseen during training. However, the environment is expected to either be seen in training or adapted to via fine tuning (Sec \ref{sec:ft}).
We conceptualize this task affordance prediction as two related aims: first, hypothesize the environmental context induced by the image, and second, predict the region in which someone may go within this hypothesized environment to perform the given task. We refer to this region of free space where a task should take place as the task's ``spatial affordance,'' and the end-to-end prediction of a task's relative region given a single egocentric image as simultaneous localization and affordance prediction.

\begin{figure*}[th]
     \centering
    \vspace*{0.5em}
    \includegraphics[width=0.9\textwidth]{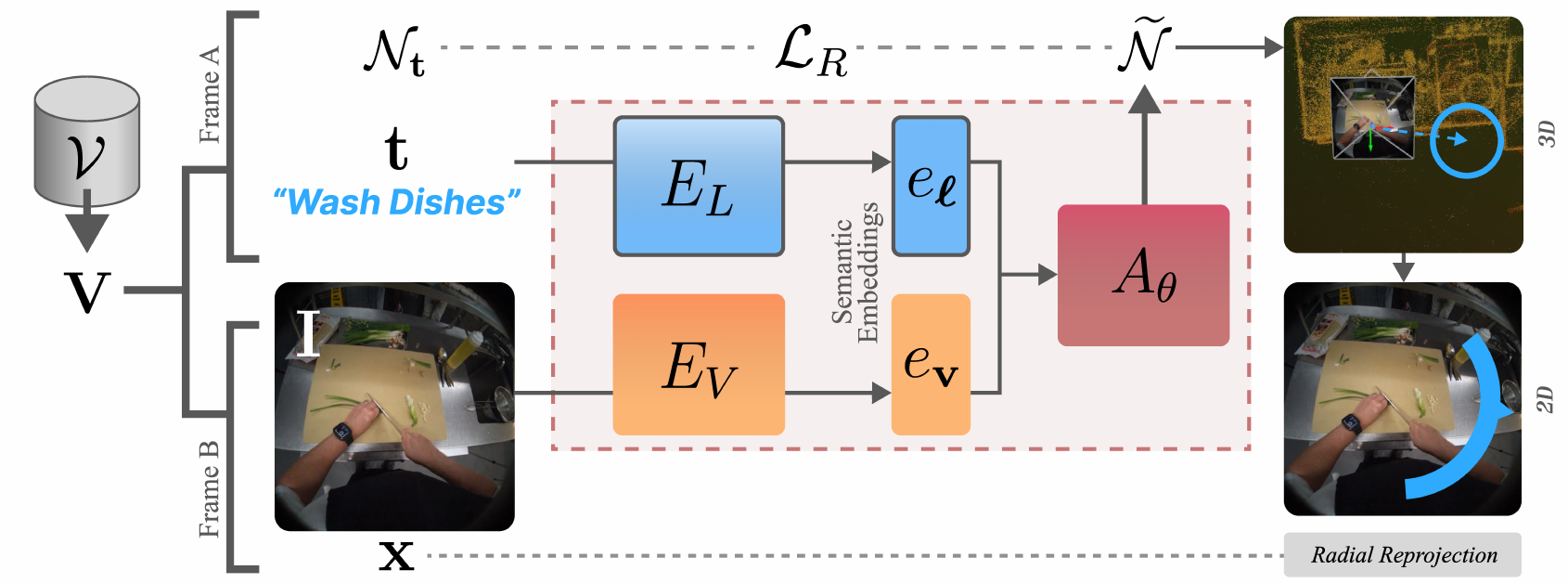}
     \caption{\small{\textbf{Model Architecture.} Given video demonstration, $\mathbf{V}$, of an activity containing several tasks, our model is trained over pairs of tasks and images selected from different times in the video. For example, a task at frame ``A'' is encoded via a (frozen) pretrained language model $E_L$ and combined with an encoding of an image from frame ``B''. Images are encoded with a pretrained vision model $E_V$ (unfrozen). This pair of encodings is finally passed to an affordance network $A_\theta$ which predicts an region where task ``B'' should take place relative to frame ``A''. The loss function $\mathcal{L}_R$ rectifies this position and compares it to the ground truth global position from task ``A''.}}
    \label{fig:arch}
     \label{fig:architecture}
\end{figure*}

\subsection{Problem Formulation}
\label{sec:problem}
Formally, given a first-person image $\mathbf{I}$ and a natural language task query $\mathbf{q}$, we would like to predict a distribution of positions where the task is performed, $\widetilde{\mathcal{D}}$, that matches the true human distribution, $\mathcal{D}$, in environment $\mathcal{E}$:
\begin{equation}
    \widetilde{\mathcal{D}}(\mathbf{q},\mathbf{I})=T(\mathcal{D}(\mathbf{q},\mathcal{E}))\label{eq:problem}
\end{equation}
Importantly, because the image $\mathbf{I}$ is egocentric, each image carries with it an implied location within the camera's environment, explicitly represented by the transform $T$.

\textbf{{Data Assumptions}}
We assume we have a dataset of videos $\mathcal{V}$ over environments $\mathcal{E}$; each video comprised of a collection of images $\mathcal{I}$, with corresponding tasks $\mathcal{T}$, and the corresponding pose where each image was observed $\mathcal{X}$. We refer to the associated collection as an annotated, localized egocentric video $\mathbf{V} \in \mathcal{V}$. Formally, $\mathbf{V} \equiv (\mathcal{X}, \mathcal{T}, \mathcal{I}),$
where elements of each set $\mathcal{X}$, $\mathcal{I}$, $\mathcal{T}$ are indexed by a frame $i$ linking the three sets together in time. In practice, such a dataset could be defined with narrated demonstrations from a subject wearing a first person camera with $\mathcal{I}$ consisting of images from the camera, $\mathcal{T}$ consisting of tasks gathered from the self-narration, and $\mathcal{X}$ consisting of poses determined by a post-collection reconstruction process such as SLAM~\cite{slam}.

\subsection{Model Architecture}
\label{sec:arch}

We model the affordance prediction task with an encoder-decoder style deep neural network architecture, first encoding the egocentric image as a vector capturing the image's semantics, and then convert this encoding into a task-conditioned prediction of the given task's performance (Figure~\ref{fig:arch}).

\noindent\textbf{Environmental Context and Image Localization}
To encode the egocentric image at the robot's current viewpoint $\mathbf{I}$, we can use pre-trained, foundational image models that have demonstrated a strong ability to capture the image's semantic information. However, such image encoding models typically capture the semantics of what is being viewed in the image rather than capture information about what to expect of the (unseen) scene surrounding the image. To address this, we fine-tune the weights of the image model, which is intended to capture the expected context given the image. The image encoder, $E_V$, is a large foundation model pre-trained on a large variety of images (such as CLIP~\cite{clip}), and will be fine-tuned on a dataset of images related to spatial affordance prediction contained in the video dataset $\mathcal{V}$. That is
\begin{equation}
e_\mathbf{v} = E_V(\mathbf{I}).
\end{equation}

\noindent\textbf{Task Encoding}
Unlike images, which need additional learned context, tasks can be encoded directly with pretrained language models such as CLIP~\cite{clip}. A task query is tokenized then encoded as a vector with a frozen, pre-trained language encoder $E_L$:
\begin{align}
  e_{\boldsymbol{\ell}} = E_L(\mathbf{q})
\end{align}
Unlike $E_V$, $E_L$ is frozen during training, as learning the context on just the image information allows the network to learn environmental context separate from downstream language task queries.

\noindent\textbf{Affordance Prediction}
Because a person may naturally move around as they accomplish a given task, each task may have a small range of positions where it was seen accomplished. 
We therefore represent the observed distribution of a task as being a normal distribution:
\begin{equation}
    \mathcal{N}_\mathbf{t}(\mu_\mathbf{t},\Sigma_\mathbf{t}) \approx \mathcal{D}(\mathbf{t}) 
\end{equation}
capturing the likely location for a person for that task across all the frames the task occurs.

The encoding vectors $e_\mathbf{v}$ and $e_\mathbf{\ell}$ represent what is expected to be around the viewer, and what the goal task is, respectively. Taken together, this should provide sufficient information for spatial affordance prediction. An affordance prediction network, $A_\theta$, is trained which takes as input these encoding vectors and produces a final 3D task region:
\begin{equation}
    A_\theta(e_\mathbf{v}, e_{\boldsymbol{\ell}}) \coloneq \widetilde{\mathcal{N}}(\mu_{\theta}, \Sigma_{\theta})
\end{equation}
whose mean is a 3D position and with 2D uncertainty constrained to lie along the ground plane with zero covariance (isotropic).
The model parameters $\theta$ are learned across environments and activities.

\subsection{Loss Function}
\label{sec:loss}
We can directly optimize Equation~\ref{eq:problem} by minimizing the difference in distributions. To ensure the predicted regions $\widetilde{N}$ are metrically meaningful, we use the Fréchet Distance, $d_{F}$, between the predicted distribution and the canonical target task distribution. Because affordance predictions happen in an egocentric frame, the target task region must be rectified before the distance loss function can be computed. We align the target task in the coordinate frame of the query image through the transform $R_{\mathbf{x}}$, and compute the error over all image-task pairs as follows:
\begin{equation}
    \mathcal{L}_{R} = \sum_{\mathbf{x}, \mathbf{I}\in\mathbf{V}}{\sum_{{\mathbf{t}}\in\mathcal{T}}}d_{F}(R_{\mathbf{x}}(\mathcal{N}_{\mathbf{t}}),\widetilde{\mathcal{N}}),\label{eq:loss}
\end{equation}computed over all videos $\mathcal{V}$. The training scheme is shown alongside the architecture in Figure~\ref{fig:arch}.

\section{Experimental Setup}\label{sec:experimental}

\subsection{Training}

\begin{table}{}
    \vspace*{0.5em}
    \centering
    \begin{tabular}{lccc}
    \toprule
    \textbf{Kitchen} & \textbf{Activity} & \textbf{Time} & \textbf{Tasks} \\
    \midrule
    \rowcolor{gray!10} 
    FAIR     & Noodles        & 9 min  & 34 \\
    GTech                       & Noodles        & 20 min & 59 \\
    \rowcolor{gray!10} 
    IIIT-H-A & Omelette       & 3 min  & 47 \\
    IIIT-H-B                    & Tomato Salad   & 2 min  & 19 \\
    \rowcolor{gray!10} 
    IndianaU & Asian Salad    & 13 min & 52 \\
    UMN-A    & Scrambled Eggs & 10 min & 33 \\
    \rowcolor{gray!10} 
    UMN-B                       & Scrambled Eggs & 6 min  & 22 \\
    SFU-A    & Scrambled Eggs & 7 min  & 16 \\
    \rowcolor{gray!10} 
    SFU-B                       & Coffee Latte   & 4 min  & 14 \\
    UAndes                      & Omelette       & 15 min & 23 \\
    \rowcolor{gray!10} 
    UPenn                       & Tomato Salad   & 8 min  & 37 \\
    UTokyo   & Omelette       & 14 min & 66 \\
    \midrule
    12 Unique & 6 Unique & 110 min & 422\\
    \bottomrule\\
    \end{tabular}
    \caption{Kitchen Activities and Tasks}
    \label{tab:kitchen_activities}
\end{table}

\label{sec:results_training}
We curated a dataset consisting of egocentric videos of people accomplishing cooking tasks from the EgoExo4D dataset~\cite{egoexo4d}, where each task is a keystep from a larger cooking activity. For example, the activity ``Making Noodles'' includes tasks such as ``Wipe hands with a kitchen towel'' and ``Add soy sauce to the noodles in the skillet.'' The resulting dataset contains nearly two hours of localized video recordings gathered from across 12 unique kitchens for a total of 422 different instances of task/environment combinations (Table~\ref{tab:kitchen_activities}).
An LLM (GPT-4~\cite{gpt4}) was used during training to augment each task description with several rephrasings which preserve the meaning of the original task. When computing keysteps for training we only consider frames where the camera has a velocity below 0.1 m/s. To stabilize our predictions in our egocentric coordinate frame, we also correct for pitch and roll of the camera.

For the pretrained vision and language encoding networks, $E_V$ and $E_L$, we used pretrained CLIP~\cite{clip} as it has been shown successful in a wide variety of language tasks. The affordance predictor network $A_\theta$ is a 4-layer MLP with 1M trainable parameters, each with layer normalization.

We randomly split the dataset into training and testing tasks (80\%/20\%), and a training and testing image set (consecutive 10\% held out), and train all models on a single V100 GPU and 10 CPU cores. Our base model was trained for 150 epochs in 7 hours of training. Our fine-tuned models are trained on all image frames of the single scene for 25 epochs, taking less than half an hour of training. The resulting models are tested in these environments on the unseen tasks.

\subsection{Baseline: Whole Scene VLMs}
\label{sec:baseline}
Similar to our proposed approach, closely related work such as CLIP-Fields~\cite{clipfields}, VLMaps~\cite{vlmaps}, and 3D-LLM~\cite{3dllm}, all build on CLIP encodings to represent semantic image information. However, unlike our proposed approach, these prior works require access to the entire 3D model of the scene at inference time. As a proxy for these types of whole-scene affordance prediction techniques, we introduce a baseline nearest-neighbor based approach which leverages pretrained CLIP as a task-similarity measure that can be applied over all images captured per scene in the dataset (no test/train split). This baseline approach, referred to as CLIP-NN, takes a CLIP text encoding of the task description $\mathbf{q}$, and a CLIP image encoding of every image in the scene $\mathbf{V}$. We can predict the best fitting image as the frame $c$ for which the cosine encoding similarity between the egocentric image and the task description text is maximized. The task position prediction is then $\mathbf{x}_{c}$, the corresponding position of the viewer at time~$c$. That is, we predict the location where the view best matches the task as evaluated by the CLIP encoding similarity. To compute the region uncertainty, we compute per-task uncertainty from all task positions, and average over all tasks in~$\mathbf{V}$.

\section{Results}\label{sec:results}
\subsection{Affordance Grounding}
\label{sec:grounding}
An immediate limitation of the baseline, and similar approaches based directly on CLIP descriptions, is that CLIP only captures the content of the image itself, rather than information about the kinds of tasks and activities that the scene affords.
This affordance grounding capability can be directly measured through a multiple-choice paradigm, where the model is used to predict which of three randomly selected task queries is most likely to take place at a given image, either the highest CLIP similarity for the baseline, or the lowest predicted distance for our method. 
Because this task does not involve any spatial prediction or out-of-view tasks, it focuses just on the model's understanding of the connection between task descriptions and an image's affordance.

\begin{figure}[t]{}
    \includegraphics[width=0.9\columnwidth]{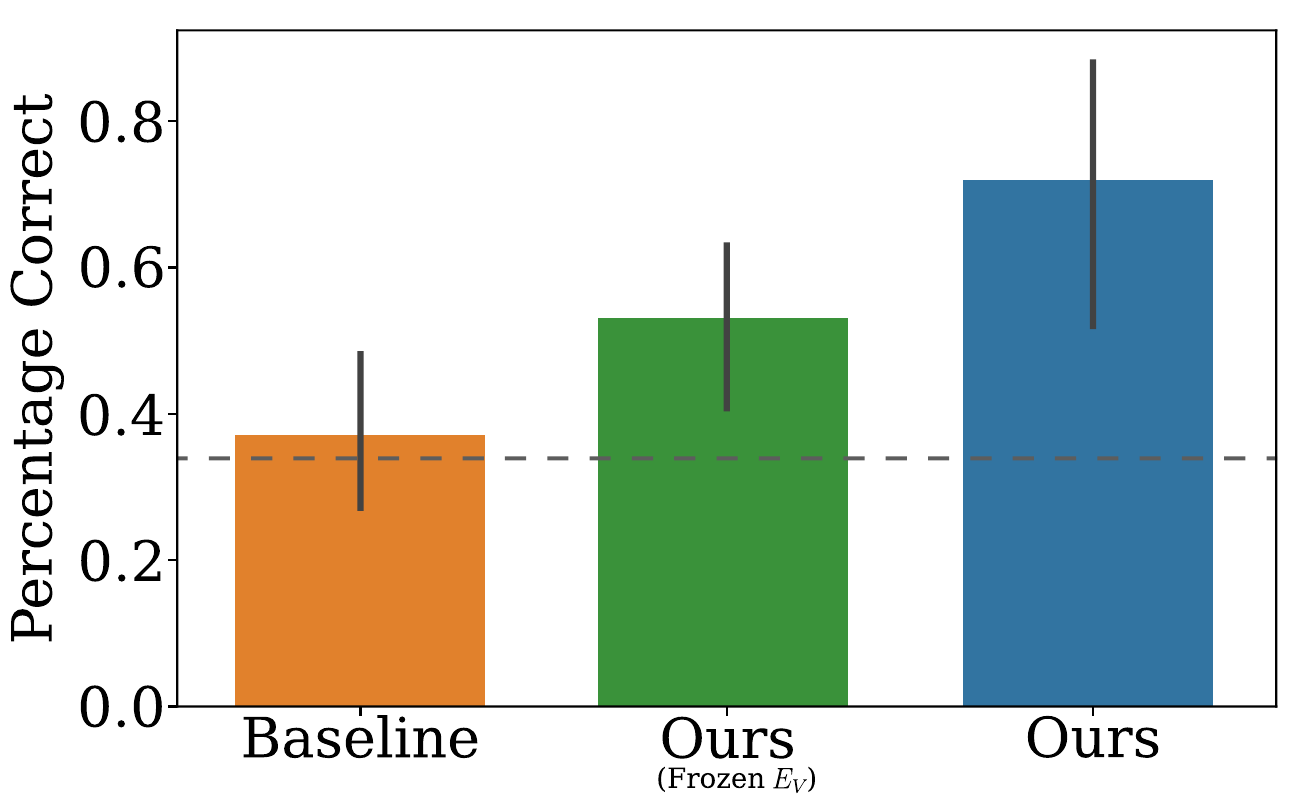}
        \caption{\small{\textbf{Affordance grounding.} When predicting from which in a set of three tasks is the most likely for a given image, the baseline (orange) performs similarly to random guessing (dashed line). Our models with a frozen language encoder (green) significantly outperforms the baseline, and our model with an unfrozen encoder (blue) nearly doubles the baseline.}}
    \label{fig:affordance}
\end{figure}

The CLIP-NN  baseline only does slightly better than random guessing (37\%), while our model has nearly double the performance of the baseline (63\%) as seen in Figure~\ref{fig:affordance}. We hypothesize this is due to CLIP encodings capturing the content of the image, rather than the activities afforded by the scene viewed from the image. Our model's ability to capture affordances comes in part by fine-tuning the vision encoder $E_V$. Retraining with a frozen $E_V$ (Original CLIP), the model still outperforms the baseline, but by a less significant margin.

\begin{table}[ht]
    \vspace*{0.5em}
    \centering
    \begin{tabular}{lcccc}
    \toprule
    \textbf{Kitchen} & 
    \textbf{Baseline} $d_{F}\downarrow$ & 
    \textbf{Ours ($E_{V \text{frozen}}$)} $d_{F}\downarrow$&
    \textbf{Ours} $d_{F}\downarrow$ \\
    \midrule
    \rowcolor{gray!10} 
    FAIR & 0.44 ± 0.4 & 0.34 ± 0.1 & \textbf{0.27 ± 0.1}\\
    
    GTech & 1.56 ± 0.9 & 0.52 ± 0.4 & \textbf{0.41 ± 0.3}\\
    
    \rowcolor{gray!10} 
    IIIT-H-A & 0.44 ± 0.2 & 0.25 ± 0.1 & \textbf{0.21 ± 0.1}\\
    
    IIIT-H-B &  1.11 ± 1.2 & 0.54 ± 0.7 & \textbf{0.52 ± 0.6}\\
    
    \rowcolor{gray!10} 
    IndianaU & \textbf{0.44 ± 0.3} & 0.56 ± 0.4 & 0.46 ± 0.3\\
    
    UMN-A & 0.55 ± 0.4 & 0.39 ± 0.2 & \textbf{0.38 ± 0.2}\\
    
    \rowcolor{gray!10} 
    UMN-B & 0.51 ± 0.5  & 0.48 ± 0.4 & \textbf{0.40 ± 0.4}\\
    
    SFU-A & 0.59 ± 0.3 & 0.34 ± 0.2 & \textbf{0.25 ± 0.2}\\
    \rowcolor{gray!10} 
    SFU-B & 0.84 ± 0.2 & \textbf{0.47 ± 0.2} & 0.52 ± 0.4\\
    
    UAndes & 0.70 ± 0.5 & 0.68 ± 0.4& \textbf{0.67 ± 0.5}\\
    
    \rowcolor{gray!10} 
    UPenn  & 0.44 ± 0.2   & 0.35 ± 0.3 &\textbf{0.28 ± 0.3}\\
    
    UTokyo & 0.44 ± 0.5       & 0.24 ± 0.2 & \textbf{0.18 ± 0.1}\\
    \midrule
    \textit{Avg Err:} & 0.68 ± 0.6  &  0.42 ± 0.3 &\textbf{0.36 ± 0.3}\\
    \bottomrule
    \end{tabular}
    \caption{\small{Task Region Localization Error, $d_{F} (m)$}}
    \label{tab:individual_result}
\end{table}

\begin{figure}[t]{}
    \vspace*{0.5em}    \includegraphics[width=0.9\columnwidth]{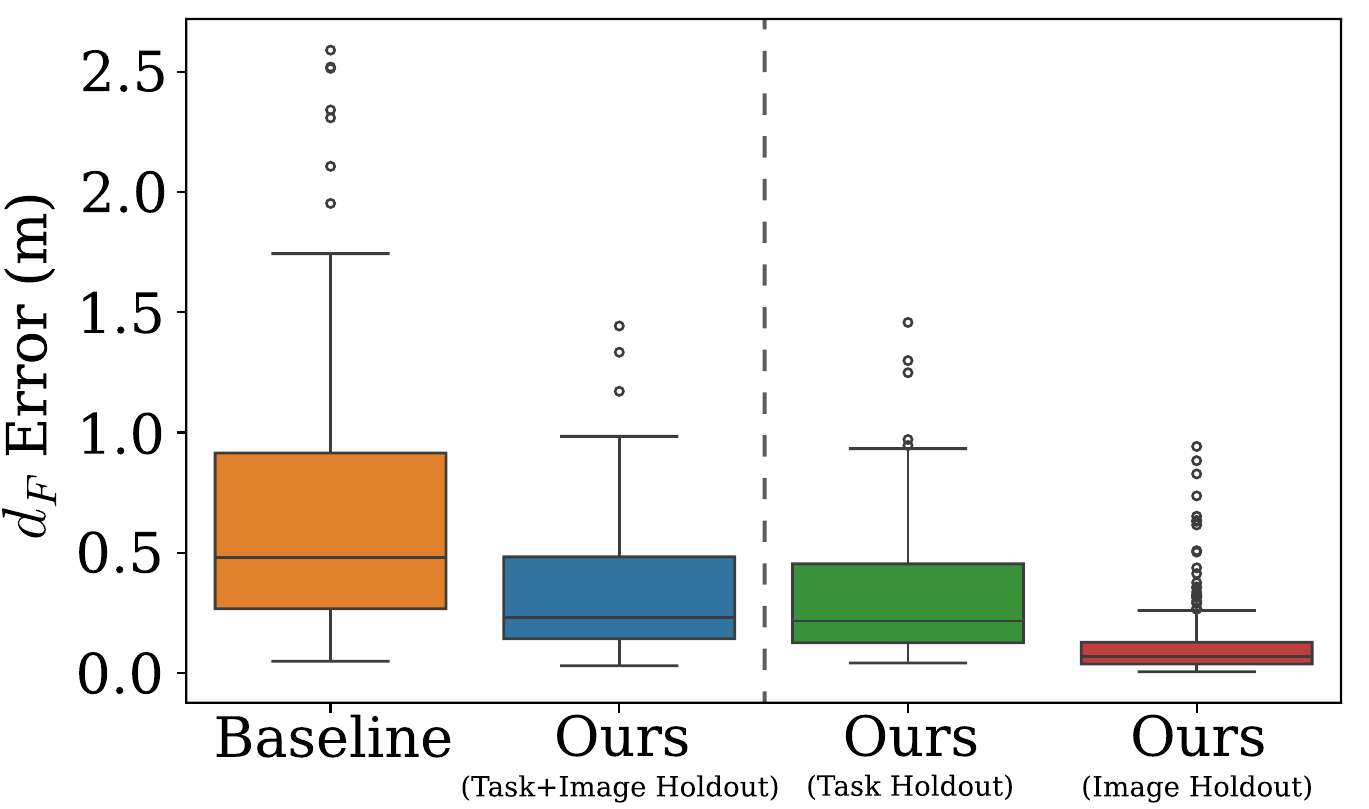}
    \caption{\small{\textbf{Task localization error.} When predicting the region of a given task, the baseline approach does well in some cases (median error of 0.48m) but has many cases with high error or significant outliers. Our approach has significantly lower error than the baseline when testing on unseen images (red), unseen tasks (green), or both unseen images and tasks (blue).}}
    \label{fig:task_loc}
\end{figure}

\subsection{Task Localization}
\label{sec:task_loc}
When compared to the baseline, our approach is also significantly more accurate at predicting where a given task will take place relative to an arbitrary egocentric viewpoint (Table~\ref{tab:individual_result}). Task localization allows the model to interface with egocentric navigation techniques such as PointNav~\cite{pointnav} and other egocentric robotic works~\cite{hutterwalk}, allowing the robot to accomplish tasks such as moving to where you need to ``heat the food.'' Our approach shows statistically significant gain over the baseline [t(82) = 4.683, p \textless 0.001] (Figure~\ref{fig:task_loc} left of dashed line) even when testing on both unseen tasks from held-out viewpoints. 

The right side Figure~\ref{fig:task_loc} shows two additional breakdowns of the task localization results tested on either only known images or known tasks. When tested on seen images and unseen tasks, the performance is nearly the same.  When tested on seen tasks and unseen images, our model has almost no error. Taken together, these two results demonstrate the quality of our model's ability to localize within a scene.

\subsection{Rephrasing Stability}
\label{sec:rephrase}
Because queries to our model arrive as natural language, the model must be able to make valid predictions across different phrasings of the same task (e.g., ``heat the skillet'' and ``warm up the pan'' should have the same predicted position). As our model was trained on a variety of rephrasings, we can expect it to handle this language variation at test time as well.
\begin{table}[b!]
\centering
\begin{tabular}{lcc}
   \toprule
     LLM & Baseline & Ours \\ 
   \midrule
    GPT-4  & 0.19m & \textbf{0.12m} \\
    LLAMA 3 & 0.18m  & \textbf{0.13m} \\
    Gemma 1.1 & 0.23m  & \textbf{0.16m} \\
   \bottomrule  
\end{tabular}
\caption{\small{Rephrasing Stability}}
\label{tab:stability}
\end{table}
To examine the stability under rephrasing, we generated new synonymous phrases for each task in our testing set, and measure the stability of our prediction over these phrases as the average standard deviation of the predicted position for each rephrasing. We test rephrasing both with the language model used in training, and two other LLMs not seen in training (LLAMA-3 8B~\cite{touvron2023llama}, and GEMMA-1.1 2B~\cite{gemmateam2024gemma}). In all cases, our model was more stable than the baseline (Table~\ref{tab:stability}), with only a small amount of variation in predicted positions for different phrasings.

\subsection{Per Scene Fine-tuning}
\label{sec:ft}

When our model is applied on new environments substantially different from those seen in the training data the quality of the results falls to below that of the baseline. This is expected in that our approach only has access to a single image with a limited field-of-view, meaning that it is forced to guess much of the scene context based only on what is a typical kitchen layout whereas the baseline has access to ground-truth labeled data for every task of every frame in the scene.
While we expect training on larger collections of scenes similar to those in testing would somewhat improve generalizing to new scenes, in practice there is too large a degree of variety in environments to reliably produce high-quality predictions of the entire scene from a single image.

A more practical approach is to fine-tune our trained model based on short demonstrations in the new environment.
Surprisingly, only a single demonstration is needed to significantly outperform the baseline. In fact, across three different kitchens unseen in training, adding a single demonstration of several tasks from one activity halves the error on unseen tasks within the same activity as shown in Table~\ref{tab:fine-tuning_2}. 

Importantly, we find that fine-tuning only the affordance head $A$ results in nearly equal performance gain compared to fine tuning both $A$ and $E_V$. This allows fine-tuning of the deployed model to happen on commodity GPUs, as optimizing $A$ requires significantly less memory than $E_V$.

\begin{table}[h]
\centering
\begin{tabular}{lccc}
   \toprule
      &  UMN-C $d_{F}\downarrow$ &  UMN-D $d_{F}\downarrow$ & SFU-C $d_{F}\downarrow$ \\ 
   \midrule
    Baseline   & 0.28\,±\,0.3m  & 0.34\,±\,0.5m & 0.69\,±\,0.4m\\
    Ours (base) & 0.55\,±\,0.3m & 0.70\,±\,0.3m & 1.05\,±\,0.5m \\
    Ours (FT $A$)& {0.20\,±\,0.2m} & {0.31\,±\,0.4m} & {0.45\,±\,0.3m} \\
    Ours (FT all)& \textbf{0.18\,±\,0.2m} & \textbf{0.26\,±\,0.3m} & \textbf{0.39\,±\,0.3m} \\
   \bottomrule
\end{tabular} \\
\makeatletter
\def\@captype{table}
\makeatother
\caption{\small{Fine-tuning on demonstrations in new scenes.}}
\label{tab:fine-tuning_2}
\end{table}

\section{Navigation Applications}
\label{sec:navigation}

\begin{figure}[t]{}
    \vspace*{0.5em}
    \includegraphics[width=1.0\columnwidth]{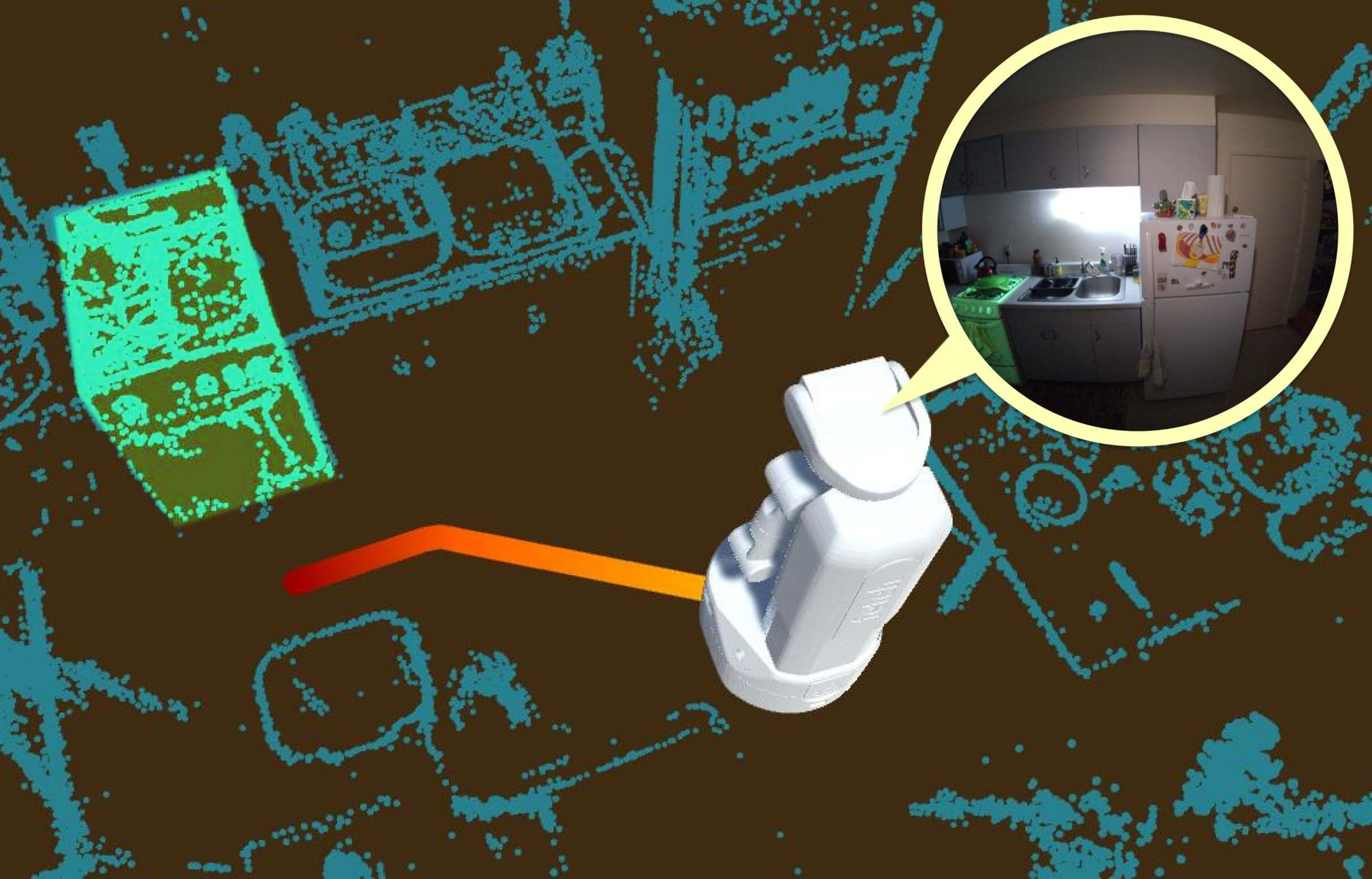}
    \caption{\small{Trajectory to ``Heat the Food" (stove highlighted).}}
    \label{fig:nav_top}
\end{figure}

\subsection{Egocentric Robot Navigation}
To characterize the ability of our system to support task-based robot navigation, we collected a new dataset of images from one of the physical environments seen in training. We then used a custom simulator to allow a robot to navigate based on these newly collected images to positions appropriate for new tasks unseen in training. We collected these images using an Aria camera~\cite{aria} as in training, and based the simulation on the Fetch robot~\cite{Wise2016FetchF} as it has similar physical affordance to humans.

An example navigation is shown in Figure~\ref{fig:nav_top}. Here a robot is given a new view (shown in the inset bubble) and asked to navigate to the task ``\textit{Heat the Food}''. Given this single egocentric robot view, the robot is able to predict the tasks' location. A navigation mesh of estimated free space is used to avoid collision during motion.

\subsection{Task Obstacles}

\begin{figure}[t!]{}
    \centering
    \vspace*{0.5em}
    \includegraphics[width=1.0\columnwidth]{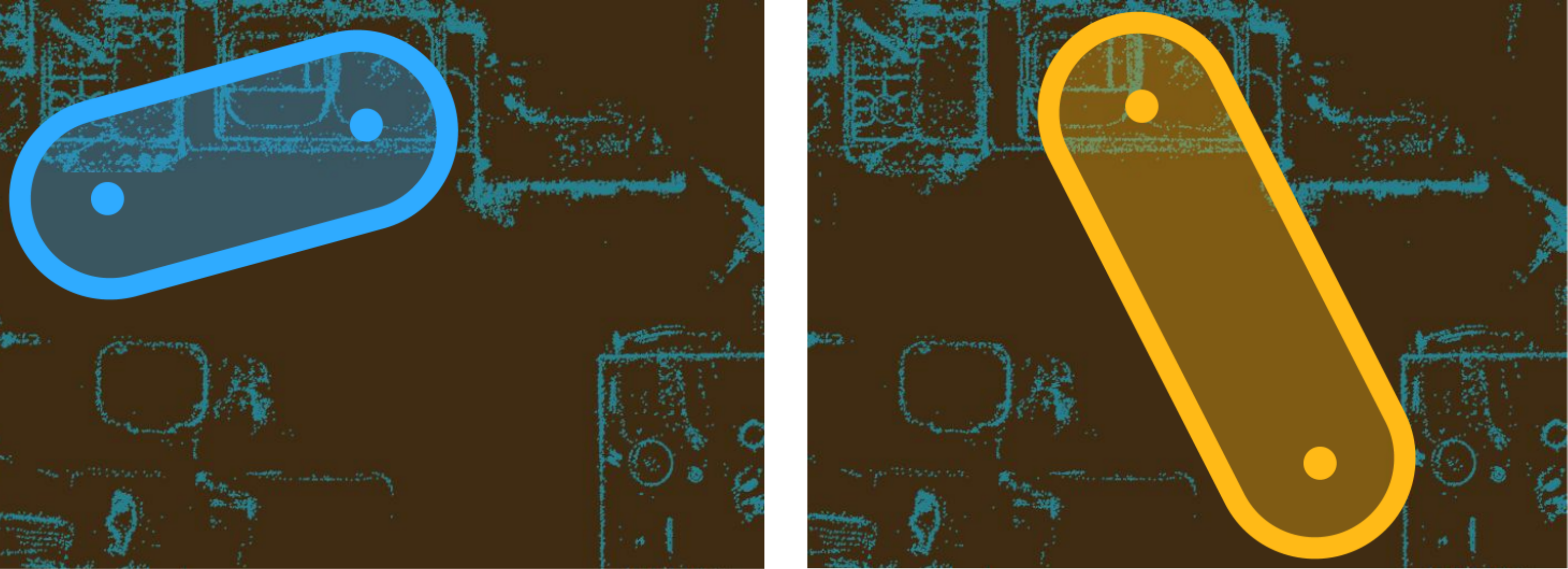}
    \caption{\small{\textbf{{Task Obstacles}} predicted across two task-sets. (left) ``Preheat the oil, and wash the dishes'' and (right) ``Get dishes from the cabinet, and serve dinner.''}}
    
    \label{fig:task_obs}
\end{figure}

In shared robot-human environments, it can be important for a robot to proactively avoid regions where a person may need to be be while doing a set of tasks. We can use the trained network to define a \textit{Task Obstacle} covering a set of locations a robot should avoid while a person is doing a set of related tasks as detailed in Algorithm~\ref{alg:taskobs}. For a given set of tasks a person may do, we first bound a safety radius of $\sigma_{\text{bound}}$ standard deviations around the predicted task regions and then encompass the entire set of bounded regions by their convex hull. The resulting task obstacle contains both the likely regions a person would be in during tasks and the areas they will likely travel between tasks, allowing a robot to plan accordingly.  Figure~\ref{fig:task_obs} shows examples of task obstacles.

\begin{algorithm}[t!]
\caption{Task Obstacle Generation }\label{alg:taskobs}
\SetAlgoLined
\DontPrintSemicolon
Load $\mathbf{A} \coloneq$ $E_V$,\: $E_L$,\: $A_{\theta}$,\:\\
Given $\mathbf{I}_{\text{current}}$,\: 
$\mathcal{T}_{\text{set}}$,\: $\sigma_{\text{bound}}$ \\
distributions $=$ $\mathbf{A}(\mathcal{T}_{\text{set}}, \mathbf{I}_{\text{current}})$\\ 
regions $=$ [\,region($\mathcal{D}$, $\sigma_{\text{bound}}$) for $\mathcal{D}$ in distributions\,]\\
points $=$ [\,discretize($\mathbf{r}$) for $\mathbf{r}$ in regions\,]\\
task\_obstacle $=$ convex\_hull(points)\\
\end{algorithm}
\vspace{-1em}

\section{Discussion}

We have introduced a new framework to predict spatial affordances of where people perform tasks within a robot's environment. Our system is trained on egocentric video demonstrations and shows generalizability to new tasks described in natural language

\textbf{Limitations \& Future Work}
Though our approach shows generalization to new tasks and novel viewpoints, this generalization is limited to scenes very similar to those seen at train time. While fine-tuning on demonstrations in the new environments helps, it still requires new training cycles which could be inconvenient in a deployed system. This limitation could be alleviated via online learning where the model is continuously updated based on live observations.
Likewise, the affordances from human demonstrations may not map one-to-one with various types of robots, and online learning or other approaches could be used to adapt between the robot and the demonstrations.
Another important limitation of our work is that all examples were taken from cooking activities in kitchens, and more environments should be considered.
Lastly, we currently assume each task region is approximated by a unimodal distribution. In the future, we would like to explore alternative forms of spatial affordance prediction, for example predicting heatmaps, or full-body poses.

{\small
	\bibliographystyle{IEEEtran}
	\bibliography{IEEEabrv,refs}
}

\end{document}